\documentclass[journal]{IEEEtran}

\ifCLASSINFOpdf
\else
\usepackage[dvips]{graphicx}
\fi
\usepackage{url}

\usepackage{graphicx}
\usepackage{amssymb}
\usepackage{multirow}

\begin{document}
	
\title{Zero-Shot Interpretable Image Steganalysis for Invertible Image Hiding}

\author{Hao Wang, Yiming Yao, Yaguang Xie, Tong Qiao, Zhidong Zhao
	\thanks{This work was supported by the Zhejiang Provincial Natural Science Foundation of China (No.LQN25F020014), the National Natural Science Foundation of China (No.62502136; No.62472135), the Zhejiang Province Key R\&D Project (No.2024C01102; No.2025C04002), the Zhejiang Provincial Natural Science Foundation of China (No.LDT23F01012F01; No.LZ23F020006), the Zhejiang Provincial Key Laboratory for Sensitive Data Security Protection and Confidentiality Management (No.2024E1004), and the Sino-French Joint Laboratory for Digital Media Forensics of Zhejiang Province. \textit{(Corresponding authors: Tong Qiao; Zhidong Zhao)}}
	\thanks{Hao Wang, Yiming Yao, Tong Qiao, and Zhidong Zhao are with the Zhejiang Provincial Key Laboratory for Sensitive Data Security Protection and Confidentiality Management, Hangzhou Dianzi University, Hangzhou 310018, China (e-mail: hwang@hdu.edu.cn; ymyao@hdu.edu.cn; tong.qiao@hdu.edu.cn; zhaozd@hdu.edu.cn).}
	\thanks{Yaguang Xie is with the Arcvideo Technology, Hangzhou 310018, China (e-mail: yxie@arcvideo.com).}
}

\markboth{Journal of \LaTeX\ Class Files, Vol. 14, No. 8, August 2015}
{Shell \MakeLowercase{\textit{et al.}}: Bare Demo of IEEEtran.cls for IEEE Journals}
\maketitle

\begin{abstract}
	Image steganalysis, which aims at detecting secret information concealed within images, has become a critical countermeasure for assessing the security of steganography methods, especially the emerging invertible image hiding approaches. However, prior studies merely classify input images into two categories (i.e., stego or cover) and typically conduct steganalysis under the constraint that training and testing data must follow similar distribution, thereby hindering their application in real-world scenarios. To overcome these shortcomings, we propose a novel interpretable image steganalysis framework tailored for invertible image hiding schemes under a challenging zero-shot setting. Specifically, we integrate image hiding, revealing, and steganalysis into a unified framework, endowing the steganalysis component with the capability to recover the secret information embedded in stego images. Additionally, we elaborate a simple yet effective residual augmentation strategy for generating stego images to further enhance the generalizability of the steganalyzer in cross-dataset and cross-architecture scenarios. Extensive experiments on benchmark datasets demonstrate that our proposed approach significantly outperforms the existing steganalysis techniques for invertible image hiding schemes.
\end{abstract}


\IEEEpeerreviewmaketitle

\section{Introduction}

\begin{figure}[ht]
	\centering
	\includegraphics[width=0.90\columnwidth]{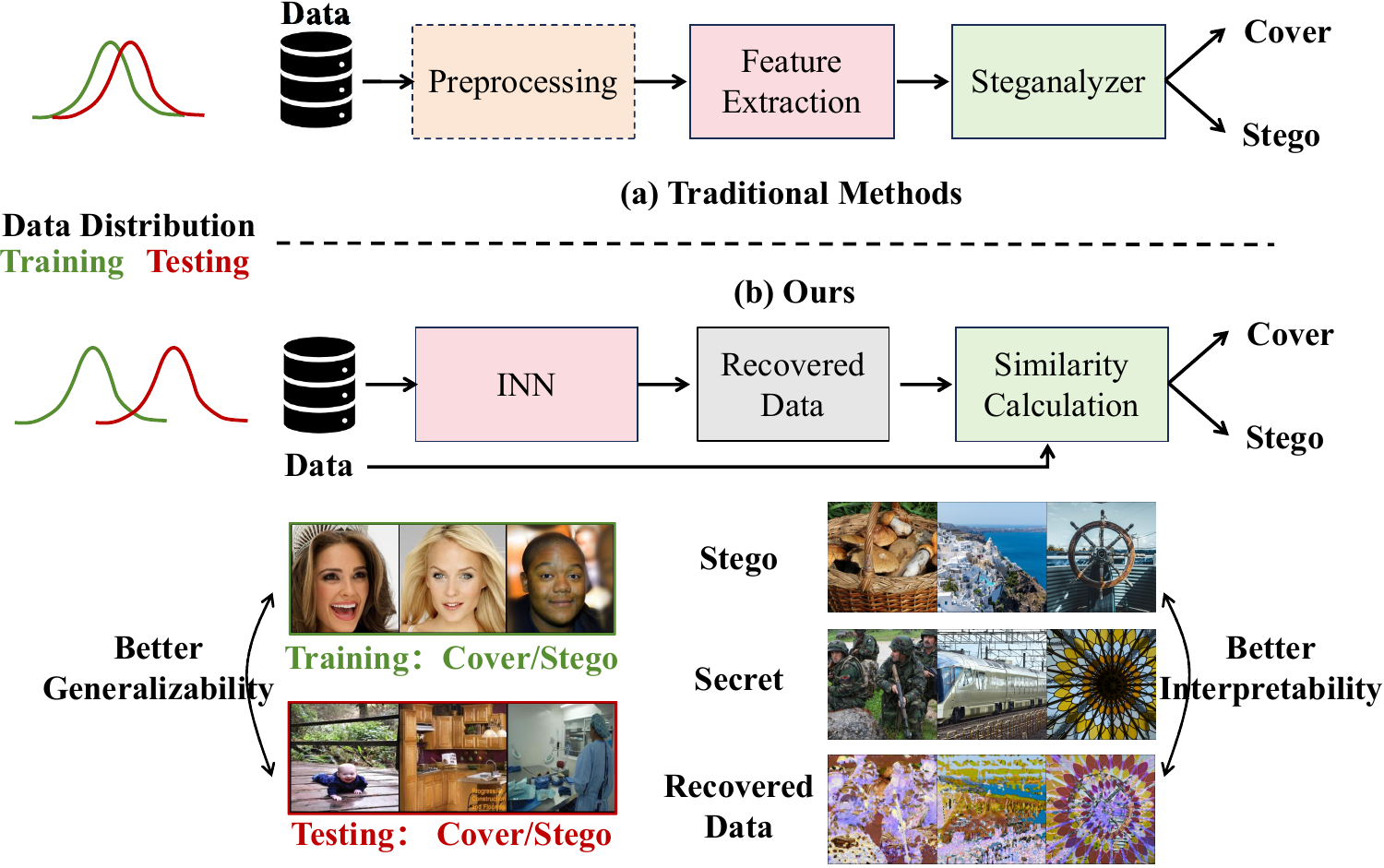}
	\caption{The motivation of our proposed method. We consider a scenario, namely zero-shot interpretable image steganalysis, to simultaneously address the drawbacks, i.e., generalizability and interpretability, of existing works.}
	\label{motivation}
\end{figure}

\IEEEPARstart{I}{mage} hiding \cite{baluja2017hiding,weng2019high,jing2021hinet,lu2021large,xu2022robust,guan2022deepmih,lan2023robust,li2024lidinet,ke2024stegformer,chen2024image,luo2025stegmamba,ye2025robust} aims to embed secret images into a cover image to generate a stego image, with key emphases on achieving large capacity and high invisibility of the hidden information. It has been widely applied in various critical fields, such as secure communication, copyright protection and digital forensics. Prior studies on image hiding can be roughly divided into two categories, i.e., autoencoder-based methods \cite{baluja2017hiding,weng2019high,ke2024stegformer,chen2024image} and invertible neural network (INN)-based approaches \cite{jing2021hinet,lu2021large,xu2022robust,guan2022deepmih,lan2023robust,li2024lidinet,luo2025stegmamba,ye2025robust}, depending on the architecture of neural networks utilized for image hiding. To the best of our knowledge, the INN-based methods generally outperform auto-encoder based approaches in balancing the quality of stego images and recovered content, thereby garnering ever-increasing attention among community.

To assess the security of INN-based image hiding methods, steganalysis techniques \cite{fridrich2012rich,denemark2014selection,tan2014stacked,xu2016structural,ye2017deep,boroumand2018deep,deng2019fast,zhang2019depth,qiao2019adaptive,you2020siamese}  were utilized to determine whether the input images contain hiding content. Early works, such as Spatial Rich Model (SRM) \cite{fridrich2012rich} and its variant maxSRMd2 \cite{denemark2014selection}, statistically computed the neighboring pixel relationships to detect steganographic schemes. With the prosperity of deep learning, researchers focused on utilizing Convolutional Neural Networks (CNNs) to enhance the accuracy of spatial domain steganalysis. For instance, SRNet \cite{boroumand2018deep} firstly introduced an end-to-end deep learning steganalysis framework, without depending on any handcrafted feature-based preprocessing.

\begin{figure*}[ht]
	\centering
	\includegraphics[scale=0.60]{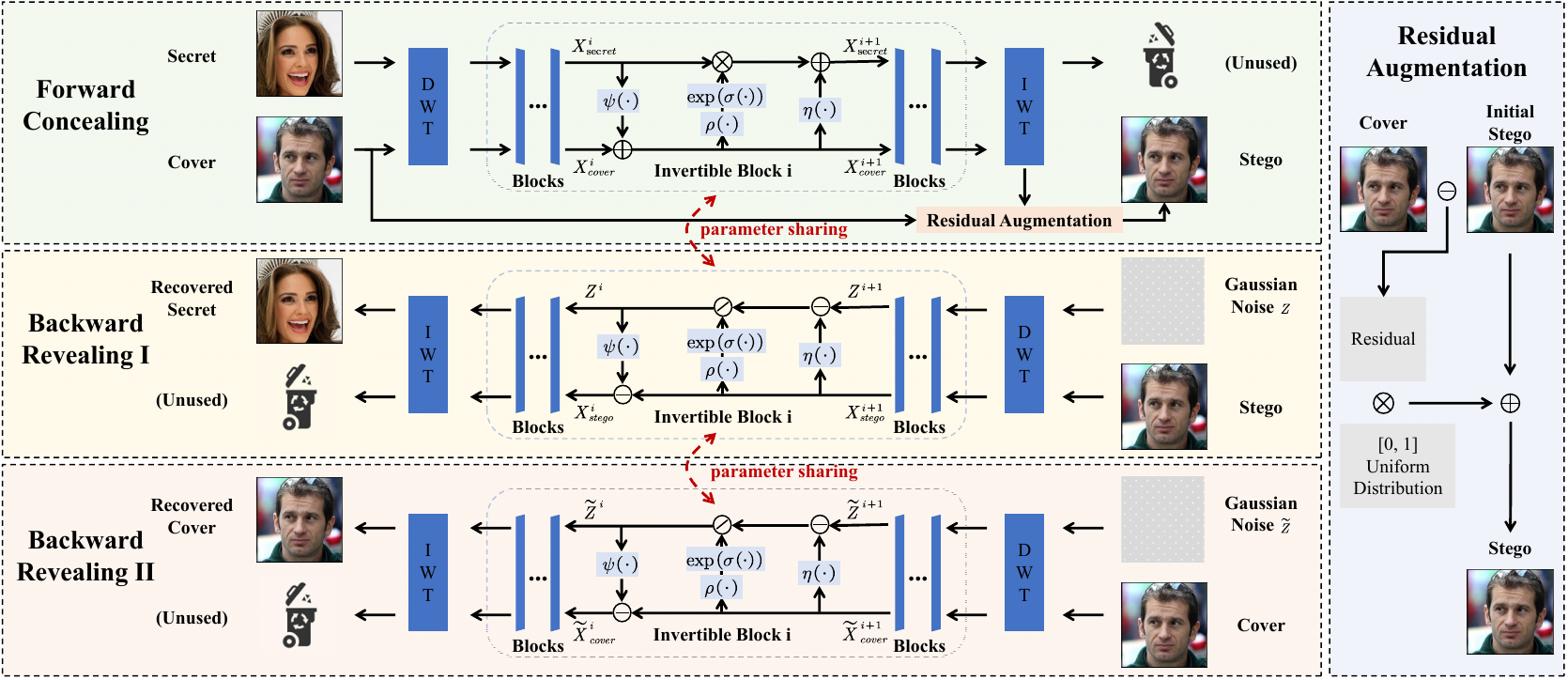}
	\caption{The framework of our proposed method, consisting of a forward concealing module and two backward revealing modules. The former embeds secret information into cover images, generating stego images on-the-fly for subsequent steganalysis. The two revealing modules function as image steganalyzers to distinguish between stego images and cover images via measuring the similarity between inputs and corresponding recovered content.}
	\label{framework}
\end{figure*}

However, two critical limitations still need to be addressed, as illustrated in Fig. \ref{motivation}. Firstly, existing steganalyzers merely estimate the probability that an input image is a stego or cover image, lacking the capability to recover the secret content concealed within the stego image and thus ultimately leading to poor interpretability of their decisions. Secondly, they also require  training of steganalysis models on datasets containing both stego and cover images, meaning that the stego images must be generated in advance. Such a constraint limits both the number of available samples and the diversity of steganographic patterns, resulting in poor generalizability of these steganalyzers in practical scenarios.

To address the aforementioned shortcomings, we propose a novel INN-based Zero-Shot Interpretable Image Steganalysis (ZSIIS) framework tailored for invertible image hiding. Specifically, to endow the INN-based framework with steganalysis capability, we reveal cover images directly from themselves, forcing our approach to learn to distinguish between cover and stego images. Moreover, to improve the generalizability of steganalyzer in cross-dataset and cross-method scenarios (i.e., challenging zero-shot setting), we design residual augmentation strategy to simulate the generation of diverse stego images. Overall, the main contributions are as follows:

\begin{itemize}
	\item We are the first to unveil the security vulnerabilities in invertible image hiding via our proposed zero-shot image steganalysis, drawing greater attention to their security.
	\item We propose a novel INN-based image steganalysis framework to simultaneously enhance the interpretability and generalizability of our proposed steganalyzer.
	\item We are the first to eliminate the requirement that training datasets must contain both cover and stego images, thereby enhancing the practical applicability.
	\item Experimental results on benchmark datasets demonstrate that our approach notably outperforms existing methods in terms of steganalysis performance.
\end{itemize}

\section{Proposed Method}
The overall structure of our proposed method is illustrated in Fig. \ref{framework}, which integrates image concealing, revealing and steganalysis into a unified framework. To enhance the practical applicability in realistic scenarios, we utilize an INN-based framework with elaborated augmentation to generate diverse stego images on-the-fly for subsequent steganalysis, avoiding labor-intensive collection and professional annotation of training dataset. Moreover, to equip the INN-based framework with steganalysis capability, we reveal cover images directly from the images themselves, thereby compelling our approach to learn the distinction between cover and stego images.

\subsection{Image Concealing with Residual Augmentation}
Firstly, we decompose both secret and cover images into low- and high-frequency wavelet sub-bands using the Discrete Wavelet Transform (DWT) block, following the approach described in HiNet \cite{jing2021hinet}. These sub-bands are then fed into a sequence of invertible blocks and ultimately converted into stego images via the Inverse Wavelet Transform (IWT) block. Concretely, the $i$-th block in the forward concealing can be formulated as follows:
\begin{equation}
	X^{i+1}_{\rm{cover}} = X^i_{\rm{cover}} + \psi(X^i_{\rm{secret}}),
\end{equation}
\begin{equation}
	X^{i+1}_{\rm{secret}} = X^i_{\rm{secret}} \odot {\rm{exp}}(\alpha(\rho(X^{i+1}_{\rm{cover}}))) + \eta(X^{i+1}_{\rm{cover}}),
\end{equation}
where $X^i_{\rm{cover}}$ and $X^i_{\rm{secret}}$ are inputs, and $X^{i+1}_{\rm{cover}}$ and $X^{i+1}_{\rm{secret}}$ are outputs, respectively. Neural networks with dense structure are utilized to represent $\psi(\cdot)$, $\rho(\cdot)$ and $\eta(\cdot)$. $\alpha(\cdot)$ is a sigmoid activation multiplied by a constant factor. Overall, the forward concealing can be described as follows:
\begin{equation}
	[\sim, X^{\rm{init}}_{\rm{stego}}] = \rm{F} (X_{\rm{secret}}, X_{\rm{cover}}),
\end{equation}
where $\rm{F}$ denotes INN and $X^{\rm{init}}_{\rm{stego}}$ means initial stego image. $\sim$ denotes the unused information (i.e., ``Unused" in Fig.\ref{framework}) generated by INN.

However, the generated stego images exhibit low diversity in steganographic patterns, owing to the lack of diverse steganography methods involved. To simulate the generation of diverse stego images, we design a simple yet effective Residual Augmentation (RA) strategy, which can be defined as follows:
\begin{equation}
	X_{\rm{stego}} = (X_{\rm{cover}} - X^{\rm{init}}_{\rm{stego}}) \odot \lambda + X^{\rm{init}}_{\rm{stego}},
\end{equation}
where $X_{\rm{stego}}$ represents the final stego image, while $\lambda$ is a coefficient sampled per initial stego image from a uniform distribution over the interval [0, 1].

\begin{table*}[ht]
	\caption{Performance comparison of zero-shot steganalysis on DIV2K. The best and second best are marked in bold and underline.}
	\centering
	\begin{tabular}{c|cccclcc}
		\hline
		\multirow{3}{*}{\textbf{Steganalyzers}} & \multicolumn{7}{c}{\textbf{Steganography Methods}}                                                                                                                                                                \\ \cline{2-8} 
		& \multicolumn{5}{c|}{Single Image Hiding}                                                                                                                    & \multicolumn{2}{c}{Multiple Images Hiding} \\ \cline{2-8} 
		& \multicolumn{1}{c|}{Weng \cite{weng2019high}}   & \multicolumn{1}{c|}{HiNet \cite{jing2021hinet}}  & \multicolumn{1}{c|}{LiDiNet \cite{li2024lidinet}} & \multicolumn{1}{c|}{StegFormer \cite{ke2024stegformer}} & \multicolumn{1}{l|}{StegMamba \cite{luo2025stegmamba}} & \multicolumn{1}{c|}{DeepMIH \cite{guan2022deepmih}}  & StegFormer \cite{ke2024stegformer}  \\ \hline
		XuNet \cite{xu2016structural}   & \multicolumn{1}{c|}{71.00}  & \multicolumn{1}{c|}{50.00}  & \multicolumn{1}{c|}{50.00}   & \multicolumn{1}{c|}{50.00}      & \multicolumn{1}{c|}{50.00}          & \multicolumn{1}{c|}{50.00}    & 50.00       \\ \hline
		YeNet \cite{ye2017deep}         & \multicolumn{1}{c|}{87.00}  & \multicolumn{1}{c|}{62.00}  & \multicolumn{1}{c|}{50.00}   & \multicolumn{1}{c|}{50.00}      & \multicolumn{1}{c|}{63.00}          & \multicolumn{1}{c|}{63.63}    & 57.50       \\ \hline
		SRNet \cite{boroumand2018deep}  & \multicolumn{1}{c|}{90.00}  & \multicolumn{1}{c|}{51.00}  & \multicolumn{1}{c|}{50.00}   & \multicolumn{1}{c|}{50.00}      & \multicolumn{1}{c|}{51.00}          & \multicolumn{1}{c|}{56.06}    & 50.00       \\ \hline
		StegNet \cite{deng2019fast}     & \multicolumn{1}{c|}{\underline{98.00}}  & \multicolumn{1}{c|}{73.00}  & \multicolumn{1}{c|}{50.00}   & \multicolumn{1}{c|}{50.00}      & \multicolumn{1}{c|}{57.00}          & \multicolumn{1}{c|}{86.36}    & 67.50       \\ \hline
		ZhuNet \cite{zhang2019depth}    & \multicolumn{1}{c|}{71.00}  & \multicolumn{1}{c|}{50.00}  & \multicolumn{1}{c|}{50.00}   & \multicolumn{1}{c|}{50.00}      & \multicolumn{1}{c|}{50.00}          & \multicolumn{1}{c|}{53.03}    & 50.00       \\ \hline
		SiaStegNet \cite{you2020siamese} & \multicolumn{1}{c|}{67.00}  & \multicolumn{1}{c|}{54.00}  & \multicolumn{1}{c|}{50.00}   & \multicolumn{1}{c|}{50.00}      & \multicolumn{1}{c|}{56.00}          & \multicolumn{1}{c|}{54.54}    & 47.50       \\ \hline
		ZSIIS (Ours)                   & \multicolumn{1}{c|}{\textbf{100.00}} & \multicolumn{1}{c|}{\textbf{100.00}} & \multicolumn{1}{c|}{\textbf{100.00}}  & \multicolumn{1}{c|}{\textbf{100.00}}     & \multicolumn{1}{c|}{\textbf{100.00}}          & \multicolumn{1}{c|}{\underline{98.48}}    & \textbf{100.00}      \\
		ZSIIS w/o RA                   & \multicolumn{1}{c|}{\textbf{100.00}} & \multicolumn{1}{c|}{\underline{78.00}} & \multicolumn{1}{c|}{\underline{52.00}}  & \multicolumn{1}{c|}{\underline{52.00}}     & \multicolumn{1}{c|}{\underline{80.00}}          & \multicolumn{1}{c|}{\textbf{100.00}}    & \underline{95.00}      \\ \hline
	\end{tabular}
	\label{sota_div2k}
\end{table*}

\subsection{Image Revealing I \& II}
For backward revealing I, stego images and noise $Z$ sampled from standard Gaussian distribution are firstly processed by DWT block to obtain low- and high-frequency wavelet sub-bands. Then they are fed into the same sequence of invertible blocks reversely and converted into recovered secret via IWT block. Specifically, the $i$-th block in the backward revealing I can be expressed as follows:
\begin{equation}
	Z^{i} = (Z^{i+1} - \eta(X^{i+1}_{\rm{stego}})) \odot {\rm{exp}}(-\alpha(\rho(X^{i+1}_{\rm{stego}}))),
\end{equation}
\begin{equation}
	X^{i}_{\rm{stego}} = X^{i+1}_{\rm{stego}} - \psi(Z^i),
\end{equation}
where $X^{i+1}_{\rm{stego}}$ and $Z^{i+1}$ are inputs, and $X^{i}_{\rm{stego}}$ and $Z^i$ are outputs, respectively. $\eta(\cdot)$, $\rho(\cdot)$, $\psi(\cdot)$ and $\alpha(\cdot)$ are the same as in the forward concealing. Overall, the backward revealing I can be formulated as follows:
\begin{equation}
	[X^{\rm{rec}}_{\rm{secret}}, \sim] = \rm{F}^{-1} (Z, X_{\rm{stego}}),
\end{equation}
where $\rm{F}^{-1}$ denotes the reversed INN and $X^{\rm{rec}}_{\rm{secret}}$ represents the recovered secret, respectively. 

Similarly, the $i$-th block in the backward revealing II can be defined as follows:
\begin{equation}
	\widetilde{Z}^{i} = (\widetilde{Z}^{i+1} - \eta(\widetilde{X}^{i+1}_{\rm{cover}})) \odot {\rm{exp}}(-\alpha(\rho(\widetilde{X}^{i+1}_{\rm{cover}}))),
\end{equation}
\begin{equation}
	\widetilde{X}^{i}_{\rm{cover}} = \widetilde{X}^{i+1}_{\rm{cover}} - \psi(\widetilde{Z}^i),
\end{equation}
where $\widetilde{X}^{i+1}_{\rm{cover}}$ and $\widetilde{Z}^{i+1}$ are inputs, and $\widetilde{X}^{i}_{\rm{cover}}$ and $\widetilde{Z}^i$ are outputs, respectively. $\widetilde{Z}$ is another noise sampled from the same Gaussian distribution as $Z$. Overall, the backward revealing II can be similarly described as follows:
\begin{equation}
	[X^{\rm{rec}}_{\rm{cover}}, \sim] = \rm{F}^{-1} (\widetilde{Z}, X_{\rm{cover}}).
\end{equation}

\subsection{Interpretable Image Steganalysis}
Based on the above two revealing modules, we can observe that the similarity between $X^{\rm{rec}}_{\rm{cover}}$ and $X_{\rm{cover}}$ should be higher than the similarity between $X^{\rm{rec}}_{\rm{secret}}$ and $X_{\rm{stego}}$. Hence, we propose a simple yet effective approach to calculate the probability of input $X$ being stego or cover image, which can be formulated as follows:
\begin{equation}
	[X^{\rm{rec}}, \sim] = \rm{F}^{-1} (Z, X)
\end{equation}
\begin{equation}
	\mathbb{I}( \rm{M}(X, X^{\rm{rec}} \le S_{\rm{thr}}),
\end{equation}
where $\mathbb{I}$ denotes the binary indicator and $S_{\rm{thr}}$ is the threshold to identify whether the
input being cover (i.e., $0$-th category) or stego (i.e., $1$-th category). $\rm{M}$ is the metric function to compute the similarity and we adopt widely used Peak Signal-to-Noise Ratio (PSNR) to represent it.

Different from the traditional steganalysis approaches that commonly train a classifier on datasets, our proposed method is more interpretable than them, owing to the explicitly recovery of secret content. Besides, classifier-free mechanism alleviates the problem of over-fitting, thereby enhancing the generalizability of our proposed approach.

\subsection{Loss Function}
For optimization, we minimize one hiding loss $\mathcal{L}_{\rm{hid}}$ and one low-frequency wavelet loss $\mathcal{L}_{\rm{freq}}$ during forward concealing, while minimizing two revealing losses $\mathcal{L}_{\rm{srev}}$ and $\mathcal{L}_{\rm{crev}}$ during backward revealing, which can be formulated as follows:
\begin{equation}
	\mathcal{L}_{\rm{hid}} = \rm{MSE}(X_{\rm{stego}}, X_{\rm{cover}}),
\end{equation}
\begin{equation}
	\mathcal{L}_{\rm{freq}} = \rm{MSE}(\mathcal{H}(X_{\rm{stego}})_{\rm{LL}}, \mathcal{H}(X_{\rm{cover}})_{\rm{LL}}),
\end{equation}
\begin{equation}
	\mathcal{L}_{\rm{srev}} = \rm{MSE}(X^{\rm{rec}}_{\rm{secret}}, X_{\rm{secret}}),
\end{equation}
\begin{equation}
	\mathcal{L}_{\rm{crev}} = \rm{MSE}(X^{\rm{rec}}_{\rm{cover}}, X_{\rm{cover}}),
\end{equation}
where Mean Squared Error (MSE) is adopted to measure the difference between output and corresponding target. $\mathcal{H}(\cdot)_{\rm{LL}}$ denotes the operation of extracting low-frequency wavelet sub-bands after the DWT decomposition.

The total loss $\mathcal{L}_{\rm{total}}$ is the weighted sum of above losses, which can be formulated as follows:
\begin{equation}
	\mathcal{L}_{\rm{total}} = \lambda_{1} * \mathcal{L}_{\rm{hid}} + \lambda_{2} * \mathcal{L}_{\rm{freq}} + \lambda_{3} * \mathcal{L}_{\rm{srev}} + \lambda_{4} * \mathcal{L}_{\rm{crev}},
\end{equation}
where $\lambda_{i}$ balance the contribution of each loss component.

\section{Experiments}
\subsection{Experimental Setup}
\textbf{Datasets:} We conduct training on CelebMaskA-HQ \cite{lee2020maskgan} dataset with 30,000 facial images and testing on benchmark datasets include DIV2K \cite{agustsson2017ntire}, COCO \cite{lin2014microsoft} and ImageNet \cite{deng2009imagenet} with 100, 5,000 and 50,000 samples, respectively. It is worth noting that training and testing data exhibit large distribution shifts to meet the requirement of zero-shot setting.

\begin{table*}[ht]
	\caption{Performance comparison of zero-shot steganalysis on COCO. The best and second best are marked in bold and underline.}
	\centering
	\begin{tabular}{c|cccclcc}
		\hline
		\multirow{3}{*}{\textbf{Steganalyzers}} & \multicolumn{7}{c}{\textbf{Steganography Methods}}                                                                                                                                                                \\ \cline{2-8} 
		& \multicolumn{5}{c|}{Single Image Hiding}                                                                                                                    & \multicolumn{2}{c}{Multiple Images Hiding} \\ \cline{2-8} 
		& \multicolumn{1}{c|}{Weng \cite{weng2019high}}   & \multicolumn{1}{c|}{HiNet \cite{jing2021hinet}}  & \multicolumn{1}{c|}{LiDiNet \cite{li2024lidinet}} & \multicolumn{1}{c|}{StegFormer \cite{ke2024stegformer}} & \multicolumn{1}{l|}{StegMamba \cite{luo2025stegmamba}} & \multicolumn{1}{c|}{DeepMIH \cite{guan2022deepmih}}  & StegFormer \cite{ke2024stegformer}  \\ \hline
		XuNet \cite{xu2016structural}   & \multicolumn{1}{c|}{70.68}  & \multicolumn{1}{c|}{50.38}  & \multicolumn{1}{c|}{49.28}   & \multicolumn{1}{c|}{49.52}      & \multicolumn{1}{c|}{45.20}          & \multicolumn{1}{c|}{46.15}    & 45.50       \\ \hline
		YeNet \cite{ye2017deep}         & \multicolumn{1}{c|}{74.68}  & \multicolumn{1}{c|}{59.00}  & \multicolumn{1}{c|}{\underline{53.34}}   & \multicolumn{1}{c|}{50.20}      & \multicolumn{1}{c|}{68.56}          & \multicolumn{1}{c|}{60.05}    & 44.80       \\ \hline
		SRNet \cite{boroumand2018deep}  & \multicolumn{1}{c|}{\underline{90.66}}  & \multicolumn{1}{c|}{52.84}  & \multicolumn{1}{c|}{50.10}   & \multicolumn{1}{c|}{49.98}      & \multicolumn{1}{c|}{50.94}          & \multicolumn{1}{c|}{55.07}    & 50.60       \\ \hline
		StegNet \cite{deng2019fast}     & \multicolumn{1}{c|}{88.58}  & \multicolumn{1}{c|}{61.68}  & \multicolumn{1}{c|}{50.16}   & \multicolumn{1}{c|}{50.00}      & \multicolumn{1}{c|}{79.88}          & \multicolumn{1}{c|}{74.60}    & 62.20       \\ \hline
		ZhuNet \cite{zhang2019depth}    & \multicolumn{1}{c|}{68.94}  & \multicolumn{1}{c|}{49.88}  & \multicolumn{1}{c|}{50.06}   & \multicolumn{1}{c|}{50.00}      & \multicolumn{1}{c|}{49.74}          & \multicolumn{1}{c|}{53.06}    & 49.65       \\ \hline
		SiaStegNet \cite{you2020siamese} & \multicolumn{1}{c|}{66.86}  & \multicolumn{1}{c|}{50.66}  & \multicolumn{1}{c|}{51.16}   & \multicolumn{1}{c|}{50.00}      & \multicolumn{1}{c|}{67.04}          & \multicolumn{1}{c|}{55.13}    & 52.65       \\ \hline
		ZSIIS (Ours)                   & \multicolumn{1}{c|}{90.28} & \multicolumn{1}{c|}{\textbf{90.22}} & \multicolumn{1}{c|}{\textbf{86.46}}  & \multicolumn{1}{c|}{\textbf{81.22}}     & \multicolumn{1}{c|}{\textbf{90.06}}          & \multicolumn{1}{c|}{\underline{90.25}}    & \textbf{90.80}     \\
		ZSIIS w/o RA                  & \multicolumn{1}{c|}{\textbf{98.10}} & \multicolumn{1}{c|}{\underline{73.44}} & \multicolumn{1}{c|}{50.26}  & \multicolumn{1}{c|}{\underline{50.50}}     & \multicolumn{1}{c|}{\underline{84.30}}          & \multicolumn{1}{c|}{\textbf{95.14}}    & \underline{85.85}     \\ \hline
	\end{tabular}
	\label{sota_coco}
\end{table*}

\begin{table*}[ht]
	\caption{Performance comparison of zero-shot steganalysis on ImageNet. The best and second best are marked in bold and underline.}
	\centering
	\begin{tabular}{c|cccclcc}
		\hline
		\multirow{3}{*}{\textbf{Steganalyzers}} & \multicolumn{7}{c}{\textbf{Steganography Methods}}                                                                                                                                                                \\ \cline{2-8} 
		& \multicolumn{5}{c|}{Single Image Hiding}                                                                                                                    & \multicolumn{2}{c}{Multiple Images Hiding} \\ \cline{2-8} 
		& \multicolumn{1}{c|}{Weng \cite{weng2019high}}   & \multicolumn{1}{c|}{HiNet \cite{jing2021hinet}}  & \multicolumn{1}{c|}{LiDiNet \cite{li2024lidinet}} & \multicolumn{1}{c|}{StegFormer \cite{ke2024stegformer}} & \multicolumn{1}{l|}{StegMamba \cite{luo2025stegmamba}} & \multicolumn{1}{c|}{DeepMIH \cite{guan2022deepmih}}  & StegFormer \cite{ke2024stegformer}  \\ \hline
		XuNet \cite{xu2016structural}   & \multicolumn{1}{c|}{67.95}  & \multicolumn{1}{c|}{50.47}  & \multicolumn{1}{c|}{48.64}   & \multicolumn{1}{c|}{49.07}      & \multicolumn{1}{c|}{43.26}          & \multicolumn{1}{c|}{44.40}    & 44.53      \\ \hline
		YeNet \cite{ye2017deep}         & \multicolumn{1}{c|}{69.98}  & \multicolumn{1}{c|}{58.13}  & \multicolumn{1}{c|}{\underline{53.13}}   & \multicolumn{1}{c|}{\underline{50.15}}      & \multicolumn{1}{c|}{62.80}          & \multicolumn{1}{c|}{53.93}    & 40.23      \\ \hline
		SRNet \cite{boroumand2018deep}  & \multicolumn{1}{c|}{\underline{89.11}}  & \multicolumn{1}{c|}{53.70}  & \multicolumn{1}{c|}{50.10}   & \multicolumn{1}{c|}{49.92}      & \multicolumn{1}{c|}{51.27}          & \multicolumn{1}{c|}{55.05}    & 50.40       \\ \hline
		StegNet \cite{deng2019fast}     & \multicolumn{1}{c|}{88.15}  & \multicolumn{1}{c|}{63.06}  & \multicolumn{1}{c|}{50.40}   & \multicolumn{1}{c|}{50.00}      & \multicolumn{1}{c|}{78.17}          & \multicolumn{1}{c|}{74.40}    & 62.72       \\ \hline
		ZhuNet \cite{zhang2019depth}    & \multicolumn{1}{c|}{69.37}  & \multicolumn{1}{c|}{49.83}  & \multicolumn{1}{c|}{50.09}   & \multicolumn{1}{c|}{49.96}      & \multicolumn{1}{c|}{49.54}          & \multicolumn{1}{c|}{53.00}    & 49.78      \\ \hline
		SiaStegNet \cite{you2020siamese} & \multicolumn{1}{c|}{66.81}  & \multicolumn{1}{c|}{51.27}  & \multicolumn{1}{c|}{52.00}   & \multicolumn{1}{c|}{49.98}      & \multicolumn{1}{c|}{67.16}          & \multicolumn{1}{c|}{54.38}    & 53.24       \\ \hline
		ZSIIS (Ours)                   & \multicolumn{1}{c|}{87.34} & \multicolumn{1}{c|}{\textbf{87.28}} & \multicolumn{1}{c|}{\textbf{82.71}}  & \multicolumn{1}{c|}{\textbf{76.55}}     & \multicolumn{1}{c|}{\textbf{87.12}}          & \multicolumn{1}{c|}{\underline{87.38}}    & \textbf{87.37}     \\
		ZSIIS w/o RA                   & \multicolumn{1}{c|}{\textbf{96.94}} & \multicolumn{1}{c|}{\underline{71.87}} & \multicolumn{1}{c|}{49.92}  & \multicolumn{1}{c|}{50.12}     & \multicolumn{1}{c|}{\underline{81.47}}          & \multicolumn{1}{c|}{\textbf{93.60}}    & \underline{84.48}     \\ \hline
	\end{tabular}
	\label{sota_imagenet}
\end{table*}

\begin{figure}[ht]
	\centering
	\includegraphics[width=0.96\columnwidth]{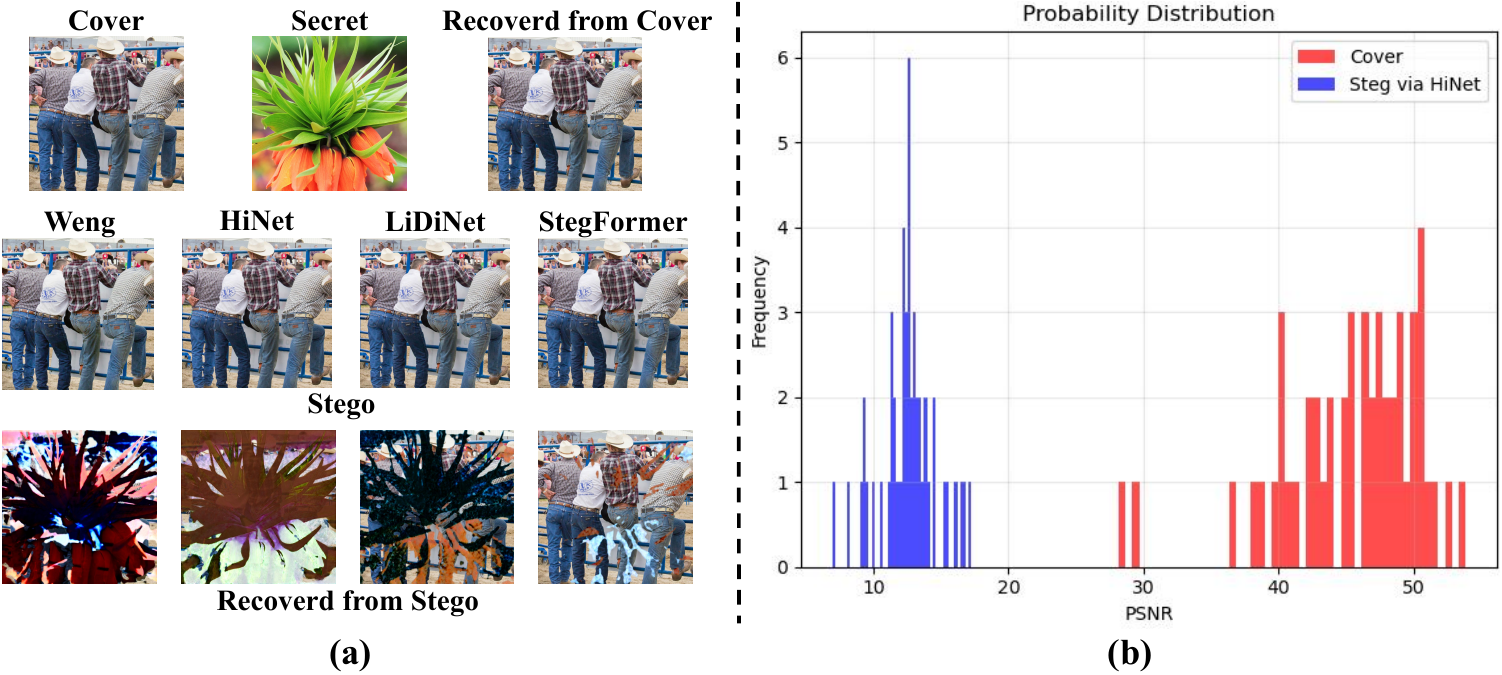}
	\caption{(a) The visualization of our proposed approach. (b) PSNR statistics between the input and its recovered data on DIV2K.}
	\label{vis}
\end{figure}

\textbf{Steganalyzers:} To evaluate the steganalysis capability, we compare our approach with six steganalyzers, XuNet \cite{xu2016structural}, YeNet \cite{ye2017deep}, SRNet \cite{boroumand2018deep}, StegNet \cite{deng2019fast}, ZhuNet \cite{zhang2019depth}, SiaStegNet \cite{you2020siamese}. The detection (i.e., cover or stego) accuracy is utilized as metric to indicate their performance.

\textbf{Steganography methods:} We select representative methods across different network paradigms: CNN-based methods (i.e., Weng \cite{weng2019high}, HiNet \cite{jing2021hinet}, LiDiNet \cite{li2024lidinet}, and DeepMIH \cite{guan2022deepmih}), Transformer-based method (i.e., StegFormer \cite{ke2024stegformer}) and Mamba-based method (i.e., StegMamba \cite{luo2025stegmamba}). For models like XuNet, YeNet, and ZhuNet, their predefined convolutional kernels are repeated across channels when processing color images.

\textbf{Implementation details:} Our approach, built upon the HiNet \cite{jing2021hinet}, is trained using randomly cropped 224$\times$224 patches from CelebMaskA-HQ \cite{lee2020maskgan}. The learning rate, batch size and the number of total epochs are $3e^{-5}$, 8 and 50. The hyper-parameters $\lambda_{1}$, $\lambda_{2}$, $\lambda_{3}$ and $\lambda_{4}$ are 1.0, 10.0, 5.0 and 5.0. At testing stage, the patch size of images after center-cropping on DIV2K, COCO and ImageNet are 1024, 256 and 256. The threshold $S_{\rm{thr}}$ is set as 25 based on the observation from Fig.\ref{vis}(b). For state-of-the-art steganalyzers, we train them to distinguish cover images from stego ones generated by our proposed approach on CelebMaskA-HQ \cite{lee2020maskgan}. The learning rate, batch size and the number of total epochs are $2e^{-4}$, 8 and 10. For steganography methods, we adopt official pre-trained models to generate stego images for subsequent detection.

\subsection{Anaysis of Experimental Results}
\textbf{Generalizability:} To evaluate the generalizability of steganalyzers, we conduct experiments under cross-dataset and cross-architecture settings. For cross-dataset setting, we observe that our proposed approach outperforms state-of-the-art steganalyzers by a significant margin across benchmark datasets, as illustrated in Tables \ref{sota_div2k}, \ref{sota_coco} and \ref{sota_imagenet}. Furthermore, state-of-the-art steganalyzers perform like random guess (i.e., 50.00\%), meaning that they fail to detect when training and testing data exhibit distribution shifts. For cross-architecture setting, our approach obtains satisfied performance (i.e., at least $76.55\%$) across steganography methods with diverse backbones.

\textbf{Interpretability:} As illustrated in Fig. \ref{vis}(a), our proposed approach can recover the hiding secret from stego, thereby improving the interpretability of detection process. For the most state-of-the-art steganography method (i.e., StegFormer), our approach can still recover the parts of hiding secret.

\textbf{Ablation studies:} From the last row of Tables \ref{sota_div2k}, \ref{sota_coco} and \ref{sota_imagenet}, we conclude that residual augmentation is crucial for improving generalizability under realistic scenarios. However, it exhibits a slight degradation when detecting stego samples generated by Weng \cite {weng2019high} and DeepMIH \cite {guan2022deepmih} in multi-image hiding scenario. We hypothesize that residual augmentation biases the model toward methods with stronger embedding strength, which may reduce sensitivity to weaker hiding schemes such as Weng and DeepMIH in the aforementioned scenarios.

\section{Conclusion}
In this letter, we propose an INN-based Zero-Shot Interpretable Image Steganalysis framework tailored for invertible image hiding, simultaneously achieving better generalizability and interpretability than existing steganalyzers. In future work, we will explore its potential in more information hiding tasks.

\end{document}